\documentclass[conference]{IEEEtran}
\usepackage[utf8]{inputenc}
\IEEEoverridecommandlockouts
\usepackage{cite}
\usepackage{float}
\usepackage{adjustbox}
\usepackage{caption}

\usepackage{url}
\usepackage[section]{placeins}
\usepackage{amsmath,amssymb,amsfonts}
\usepackage{algorithmic}
\usepackage{multirow}
\usepackage{multicol}
\usepackage{balance}
\usepackage{amsmath}
\usepackage{booktabs}
\usepackage{caption}
\usepackage{graphicx}
\usepackage{hyperref}
\hypersetup{
    colorlinks=true,      
    linkcolor=red,       
    anchorcolor=black,    
    citecolor=blue,      
    filecolor=cyan,       
    menucolor=red,        
    runcolor=cyan,        
    urlcolor=magenta      
}
\usepackage{subcaption}
\usepackage{textcomp}
\usepackage{xcolor}
\usepackage{fancyhdr}
\usepackage{tikz}
\usepackage{pifont} 

\def\BibTeX{{\rm B\kern-.05em{\sc i\kern-.025em b}\kern-.08em
    T\kern-.1667em\lower.7ex\hbox{E}\kern-.125emX}}

\begin{document}

\title{Predicting Bad Goods Risk Scores with ARIMA Time Series: A Novel Risk Assessment Approach
}

\author{\IEEEauthorblockN{Bishwajit Prasad Gond}
\IEEEauthorblockA{\textit{Department of Computer Science and Engineering,} \\
\textit{National Institute of Technology Rourkela, Odisha, India} \\
\href{https://orcid.org/0000-0003-3640-0463}{0000-0003-3640-0463}\\
}}
\maketitle
\begin{abstract}
The increasing complexity of supply chains and the rising costs associated with defective or substandard goods (“bad goods”) highlight the urgent need for advanced predictive methodologies to mitigate risks and enhance operational efficiency. This research presents a novel framework that integrates Time Series ARIMA (AutoRegressive Integrated Moving Average) models with a proprietary formula specifically designed to calculate bad goods after time series forecasting. By leveraging historical data patterns, including sales, returns, and capacity, the model forecasts potential quality failures, enabling proactive decision-making. ARIMA is employed to capture temporal trends in time series data, while the newly developed formula quantifies the likelihood and impact of defects with greater precision. Experimental results, validated on a dataset spanning 2022–2024 for Organic Beer-G 1 Liter, demonstrate that the proposed method outperforms traditional statistical models, such as Exponential Smoothing and Holt-Winters, in both prediction accuracy and risk evaluation. This study advances the field of predictive analytics by bridging time series forecasting, ARIMA, and risk management in supply chain quality control, offering a scalable and practical solution for minimizing losses due to bad goods.  
\end{abstract}


\section{INTRODUCTION}\label{sec:intro}

In modern industrial systems, identifying and mitigating defective, expired or substandard products, called ``bad goods''---such as manufacturing defects or unsold perishable items like Organic Beer-G 1 Liter poses a significant challenge. These issues lead to financial losses, reputational damage, and supply chain disruptions. Conventional methods, such as statistical process control and manual planning, are often inadequate to manage the complexity of large-scale operations~\cite{box2015time}. The rise of big data and advanced analytics has placed predictive approaches as a vital tool for proactively addressing these risks~\cite{li2020hybrid}.

This research presents a predictive analytics framework that combines Time Series ARIMA models with risk assessment to anticipate occurrence of bad goods and inform supply chain strategies. Using time-series data such as sales, returns, and production capacity, ARIMA models effectively capture temporal patterns, surpassing simpler techniques such as Exponential Smoothing in detecting linear trends~\cite{li2020hybrid}.A risk assessment layer, inspired by Failure Mode and Effects Analysis (FMEA), quantifies failure likelihood and impact, enhancing decision-making~\cite{stamatis2003failure}.

Motivated by the limitations of reactive quality management, this research aims to deliver a data-driven proactive solution for industries such as beverage production and retail. The objectives are to: (1) build a precise ARIMA-based predictive model, (2) incorporate risk assessment for practical insights, and (3) benchmark its efficacy against existing standards. By reducing bad goods, this approach promises cost savings and improved customer satisfaction across manufacturing, retail, and logistics sectors.

The remaining sections of the paper are structured as follows: Section \ref{sec:basicconcepts} provides an overview of the basic concepts. Section \ref{sec:relatedwork} reviews recent literature on predictive analytics for bad goods. Section \ref{sec:framework} presents our Time Series ARIMA with Risk Assessment framework for bad goods, focusing on preprocessing. Section \ref{sec:experiment} describes our experimental setup, including datasets used. Section \ref{sec:results} elaborates on the use of ARIMA for predictive analytics of bad goods, and result analysis. In Section \ref{sec:compare}, we compare our approach with state-of-the-art techniques in predictive analytics for bad goods. Finally, Section \ref{sec:future} concludes our work and suggests some future research directions.

\section{BASIC CONCEPTS}\label{sec:basicconcepts}
The proposed framework in this research relies on three foundational pillars: time series analysis, specifically Time Series ARIMA, risk assessment, and quality control principles. Each concept is integral to predicting the occurrence of ``bad goods''—defective or substandard products—and mitigating their impact on industrial systems, particularly for Organic Beer-G 1 Liter.

\textbf{Time-Series Analysis (ARIMA)}: Time-series analysis involves examining data points collected sequentially over time to identify trends, seasonality, and anomalies. In the context of bad goods, time series data may include production metrics, sales, returns, and capacity, exhibiting temporal patterns indicative of quality issues. The autoregressive integrated moving average model (ARIMA), a well-established time series forecasting method, is extensively used to predict data by capturing linear relationships and ensuring the direct separation of stationarity \cite{box2015time}. ARIMA excels in its interpretability and suitability for stationary or differenced non-stationary data, though it may falter with highly non-linear or intricate patterns compared to advanced machine learning approaches \cite{li2020hybrid}.

\textbf{Risk Assessment}: Risk assessment assesses the risk likelihood and impacts in order to support prioritization of supply chain decision. In quality management, methodologies like the Failure Mode and Effects Analysis (FMEA) score defect severity, occurrence, and detectability, providing an organized approach to risk ranking \cite{stamatis2003failure}. The integration of risk scoring with ARIMA-based forecasts in this study provides a forward leap from conventional forecasting, offering proactive industry solutions to bad goods risk management. Unlike stand-alone risk models, the integrated model fills an important void by utilizing predictive analytics. 

\textbf{Quality Control}: The quality control in the supply chains is the protection of the product integrity and therefore minimization of defects and returns. For perishable products such as beer, the freshness, shelf-life, production capabilities are important in analyzing the risk of bad goods.
This study embeds quality control concepts to direct ARIMA forecasting and risk analysis to make certain that they fit to operational realities and industry norms.



Collectively, these elements underpin a predictive analytics system designed for bad goods management in Organic Beer-G 1 Liter. Time-Series ARIMA provides robust forecasting, risk assessment ensures practical relevance, and quality control enhances operational alignment. This study integrates these components to develop a cohesive model that advances quality assurance in supply chain environments.

\section{RELATED WORKS}\label{sec:relatedwork}
Using predictive analytics in quality control and defect detection for supply chain operations is a growing research area. Traditional time series forecasting methods, such as the Autoregressive Integrated Moving Average (ARIMA) model, have been widely used for time series forecasting in industrial applications \cite{box2015time}. However, ARIMA is designed for linear and stationary data, and it is not suitable for non-linear, high-dimensional, or noisy data which is characteristic of modern supply chains, leading to alternate solution methodologies \cite{li2020hybrid}. Machine learning algorithms, such as Support Vector Machines (SVM) and Random Forests, have been applied to anomaly detection and defect prediction in manufacturing, but they often cannot model long-term temporal dependencies necessary for bad goods prediction~\cite{Aljohani2023Predictive}\cite{Jawad2025Designing}.

Advanced time series methods, particularly deep learning approaches, have revolutionized forecasting in supply chains. Recurrent Neural Networks (RNNs), especially Long Short-Term Memory (LSTM) networks, have proven effective for modeling sequential data and predicting equipment failures or product defects over extended periods \cite{malhotra2015long}. Convolutional Neural Networks (CNNs) have also been adapted for time series tasks, extracting local features from multivariate inputs to improve prediction accuracy \cite{bai2018empirical}. Hybrid models combining CNNs, LSTMs and BiLSTMs have shown promise in supply chain demand forecasting, as demonstrated by \cite{li2020hybrid}\cite{Necula2025Hybrid}, yet their computational complexity and data requirements can limit scalability for real-time applications. Despite these advancements, few studies have specifically targeted bad goods prediction, leaving a gap in quality-specific predictive frameworks for supply chains.

Risk assessment, a cornerstone of supply chain management, has been integrated with predictive models to prioritize critical events. Techniques such as Failure Mode and Effects Analysis (FMEA) and probabilistic risk scoring have been employed to evaluate defect impacts and guide decision-making \cite{stamatis2003failure}. However, their integration with time series models like ARIMA remains underexplored. Existing research often treats forecasting and risk analysis as separate tasks, reducing the ability to provide unified, actionable insights for bad goods management. This study addresses this gap by proposing a novel framework that combines Time Series ARIMA with risk assessment, leveraging historical data to predict and score bad goods risks for Organic Beer-G 1 Liter. By building on state-of-the-art time series and risk management techniques, this research advances predictive analytics for quality assurance in supply chains.
\section{PROPOSED FRAMEWORK}\label{sec:framework}

\begin{figure*}[h]
  \centering
  \includegraphics [angle=0,origin=c,width=0.9\textwidth]{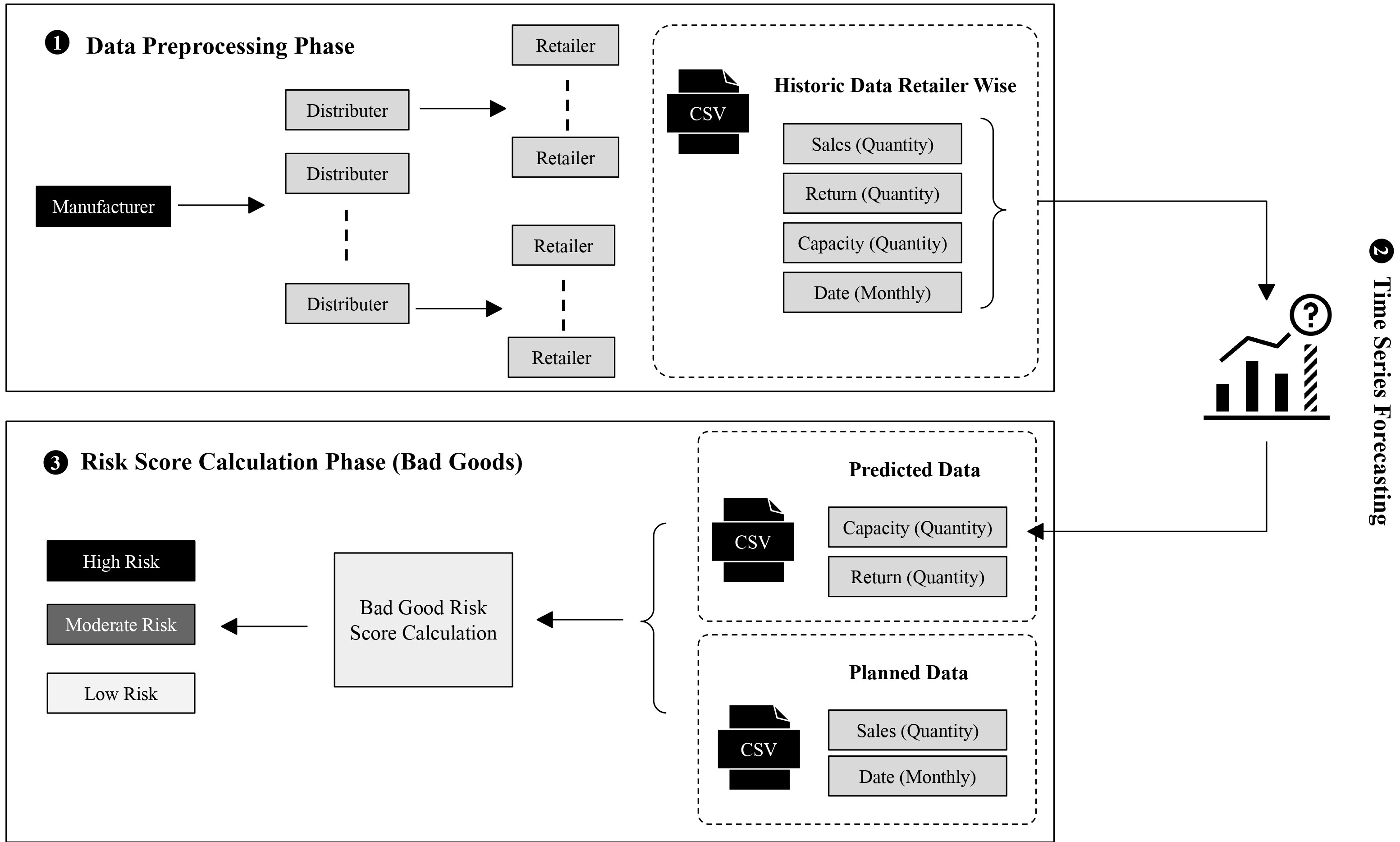}
  \caption{Proposed Architecture for Predicting Bad Goods Risk Score}
  \label{fig:arch}
\end{figure*}

\subsection{Data Preprocessing Phase}
The first phase, shown in the top section of Figure~\ref{fig:arch}, focuses on collecting and preprocessing historical data from the supply chain. Data originates from the manufacturer, which distributes products through multiple distributors to various retailers. The data flow involves:

\textbf{- Data Sources:} Historical data is gathered retailer-wise, capturing key metrics such as sales (quantity), returns (quantity), capacity (quantity), and date (monthly). These data points are critical for understanding the operational performance and return patterns of goods.

\textbf{- Data Collection:} The data is collected from manufacturer who store data at each level like distributors, who then it to retailers. This hierarchical flow ensures comprehensive coverage of the supply chain.

\textbf{- Data Storage:} The collected data is stored in CSV format, enabling structured and scalable data management. The CSV files include fields for sales, returns, capacity, and date, which are essential for subsequent time series analysis.

This phase ensures that raw data is cleaned, organized, and formatted for input into the time series forecasting model, setting the foundation for accurate risk prediction.

\subsection{Time Series Forecasting Phase}
The second phase, shown in the middle section of Figure~\ref{fig:arch}, leverages Time Series ARIMA (AutoRegressive Integrated Moving Average) models to forecast future trends in sales, returns, and capacity. This phase is critical for predicting potential bad goods risks based on historical patterns:

\textbf{- Input Data:} The preprocessed CSV data from the Data Preprocessing Phase is used as input for the ARIMA model. The model analyzes time series data (monthly sales, returns, and capacity) to identify trends, seasonality, and anomalies.

\textbf{- Forecasting Process:} ARIMA performs time series forecasting to predict future values of sales (quantity), returns (quantity), and capacity (quantity) over a specified period. The forecast is represented graphically, indicating potential fluctuations or risks in bad goods (e.g., high return rates).

\textbf{- Output:} The output of this phase is a set of predicted data, also stored in CSV format, which serves as input for the subsequent risk calculation phase. The time series forecasting helps in identifying periods of high risk for bad goods based on historical trends.

This phase bridges historical data analysis with predictive insights, enabling proactive risk management for bad goods in the supply chain.

\subsection{Risk Score Calculation Phase (Bad Goods)}
The third phase, illustrated in the bottom section of Figure~\ref{fig:arch}, focuses on calculating a bad goods risk score using the predicted data from the Time Series Forecasting Phase, combined with planned data for comparison and risk assessment:

\textbf{- Risk Levels:} The system categorizes risks into three levels—High Risk, Moderate Risk, and Low Risk—based on the predicted return quantities and sales trends. These categories help prioritize interventions for defective or returned goods.

\textbf{- Risk Score Calculation}: The bad goods risk score is computed by comparing predicted data (sales and return quantities from the Time Series Forecasting Phase) with planned data (target sales and capacity from CSV files). This comparison identifies deviations that indicate potential risks, such as excessive returns or capacity shortages, which could signal bad goods issues.

\textbf{- Output Data:} The calculated risk scores are stored in CSV format, along with predicted and planned data, for further analysis, reporting, or decision-making. The risk scores guide stakeholders (manufacturers, distributors, retailers) in mitigating bad goods risks through targeted actions.

This phase integrates predictive analytics with risk assessment to provide actionable insights, enabling stakeholders to address bad goods proactively and optimize supply chain operations.

The proposed architecture ensures a systematic approach to predicting and managing bad goods risks, leveraging the power of Time Series ARIMA for forecasting and a structured risk assessment framework for actionable outcomes. This methodology enhances decision-making across the supply chain, from manufacturers to retailers, to minimize losses and improve product quality.

\subsection*{Definitions and Formulas}
\begin{itemize}
    \item \textbf{Return Rate (\%):} Forecasted monthly for the next calendar year, based on historical patterns. This metric indicates the percentage of sold items that are returned.
    \begin{equation}
        \text{Return Rate} = \frac{\text{Actual Returns}}{\text{Sales Quantity}} \label{eq:return_rate}
    \end{equation}

    \item \textbf{Expected Return Quantity:} Represents the anticipated number of returns based on sales volume and return rate. Useful for inventory planning.
    \begin{equation}
        \text{Expected Return Qty} = \text{Sales Qty} \times \text{Return Rate} \label{eq:expected_return_qty}
    \end{equation}

    \item \textbf{Retailer Inventory Capacity (Monthly):} Forecasted and derived based on historical patterns. It measures the available inventory space after accounting for reruns.
    \begin{equation}
        \text{Inventory Capacity} = \text{Sales Qty} - \text{Rerun Qty} \label{eq:inventory_capacity}
    \end{equation}

    \item \textbf{Freshness Ratio:} Quantifies the freshness of inventory relative to its shelf life. A higher ratio indicates fresher stock.
    \begin{equation}
        \text{Freshness Ratio (FR)} = \frac{\text{Freshness}}{\text{Shelf Life}} \label{eq:freshness_ratio}
    \end{equation}

    \item \textbf{Bad Good Risk Score:} Assesses the risk of inventory spoilage or obsolescence by comparing expected returns to inventory capacity, adjusted by freshness.
    \begin{equation}
        \text{Risk Score} = \left( \frac{\text{Expected Return Quantity}}{\text{Retailer Inventory Capacity}} \right)^{\text{FR}} \label{eq:risk_score}
    \end{equation}
\end{itemize}
    



To gain an in-depth understanding of the architecture and data, one can refer to the source code of our model available on GitHub\footnote{\url{https://github.com/bishwajitprasadgond/BGRS}}.

\section{EXPERIMENTAL SETUP}\label{sec:experiment}
Our experimental setup was designed to evaluate the effectiveness of Time Series ARIMA models in predicting bad goods risk scores, integrated with risk assessment techniques for supply chain analysis. It comprises the following components:

\begin{enumerate}
    \item \textbf{Analysis Environment:}
    \begin{itemize}
        \item \textbf{Host OS:} Windows 11 running on a machine equipped with an Intel Core i5 processor, 16 GB RAM, and a 250 GB HDD.
    \end{itemize}
    
    \item \textbf{Development Environment:}
    \begin{itemize}
        \item \textbf{Python Version:} Python 3.12 used for implementing and executing the Time Series ARIMA models and data processing scripts.
        \item \textbf{IDE:} Spyder, integrated within the Anaconda distribution, utilized for coding, debugging, and running the ARIMA-based predictive analytics.
    \end{itemize}
    
    \item \textbf{Dataset:}
    \begin{itemize}
        \item A comprehensive dataset of historical supply chain data, including retailer-wise sales (quantity), returns (quantity), capacity (quantity), and date (monthly), was collected from manufacturers, distributors, and retailers. The dataset spans from January 2022 to December 2024, covering diverse product categories, with a focus on bad goods (e.g., defective or returned products) for risk score calculation.
    \end{itemize}
    
    \item \textbf{Time Series ARIMA Implementation:}
    \begin{itemize}
        \item The ARIMA model was implemented using Python libraries such as `statsmodels` and `pandas` for time series forecasting of sales, returns, and capacity. The model was tuned to optimize parameters (p, d, q) for predicting future trends and identifying potential bad goods risks based on historical patterns.
    \end{itemize}
    
    \item \textbf{Data Processing and Risk Assessment:}
    \begin{itemize}
        \item Data preprocessing was performed on the Windows 11 host to clean, organize, and format the CSV data into a suitable structure for ARIMA analysis. Risk scores were calculated by comparing ARIMA-predicted data with planned data (target sales and capacity), categorizing risks into High Risk, Moderate Risk, and Low Risk for bad goods.
    \end{itemize}
\end{enumerate}

This setup facilitated controlled experimentation to assess the performance of Time Series ARIMA in predicting bad goods risk scores and enhancing risk assessment for supply chain management. For detailed information about our malware detector's experimental setup, dataset, and source code, refer to our GitHub repository.

\section{RESULT ANALYSIS}\label{sec:results}

This section presents a detailed statistical analysis of the time series data for the Beer-G product, focusing on key variables such as retailer capacity, rate of return, bought quantity, return quantity, and their interrelationships, as illustrated in the provided figures. The analysis is grounded in the autocorrelation functions, distribution of numerical variables, historical and forecasted trends, correlation heatmaps, and additional distributional and relational visualizations, which collectively provide insights into the dynamics of Beer-G product performance over time.

\subsection{Autocorrelation Analysis}
\begin{figure}[h]
    \centering
    \includegraphics[width=8.5cm]{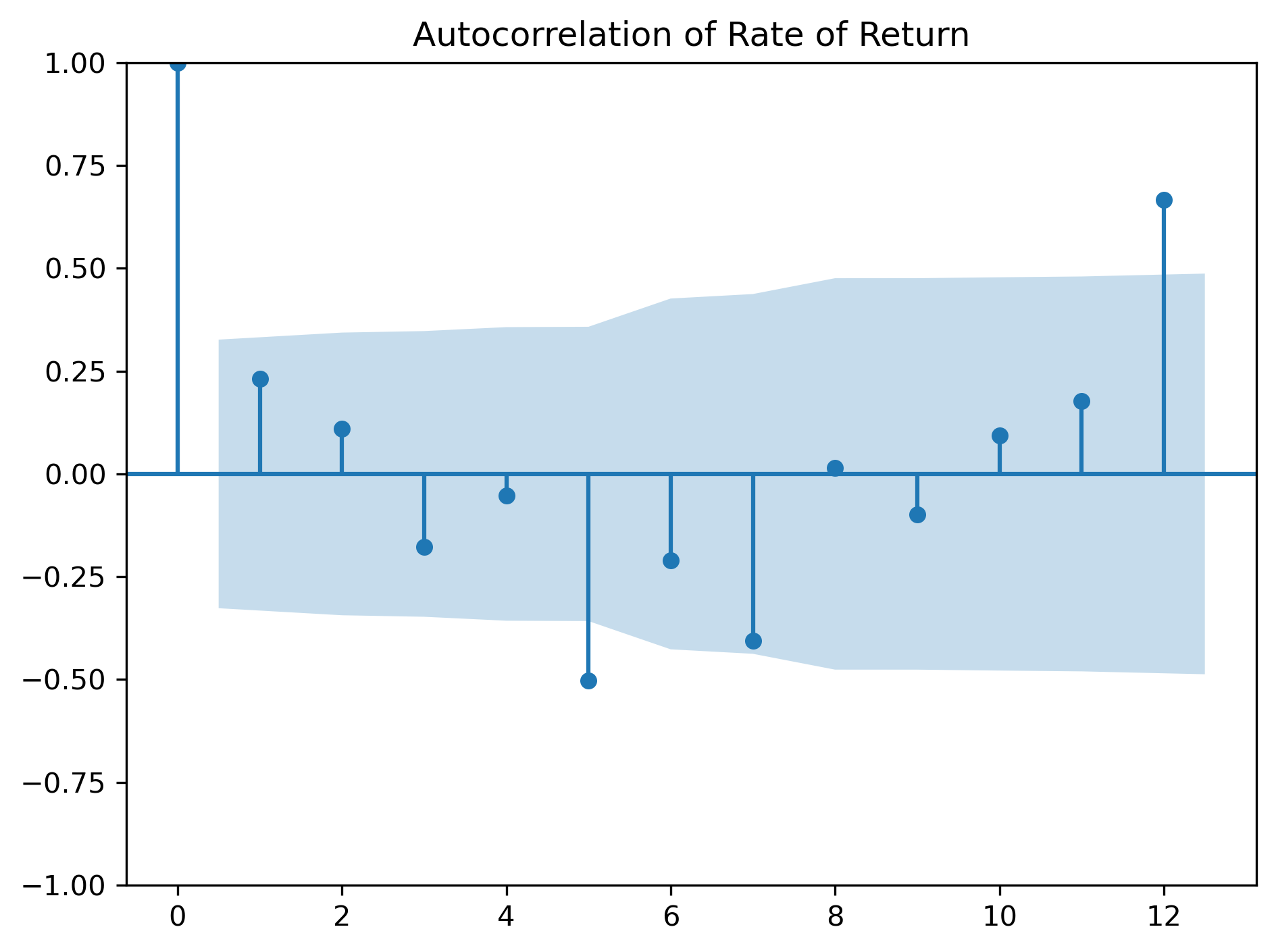}
    \caption{Autocorrelation of Retailer Capacity}
    \label{fig:autocorrelation_retailer_capacity}
\end{figure}

\begin{figure}[h]
    \centering
    \includegraphics[width=8.5cm]{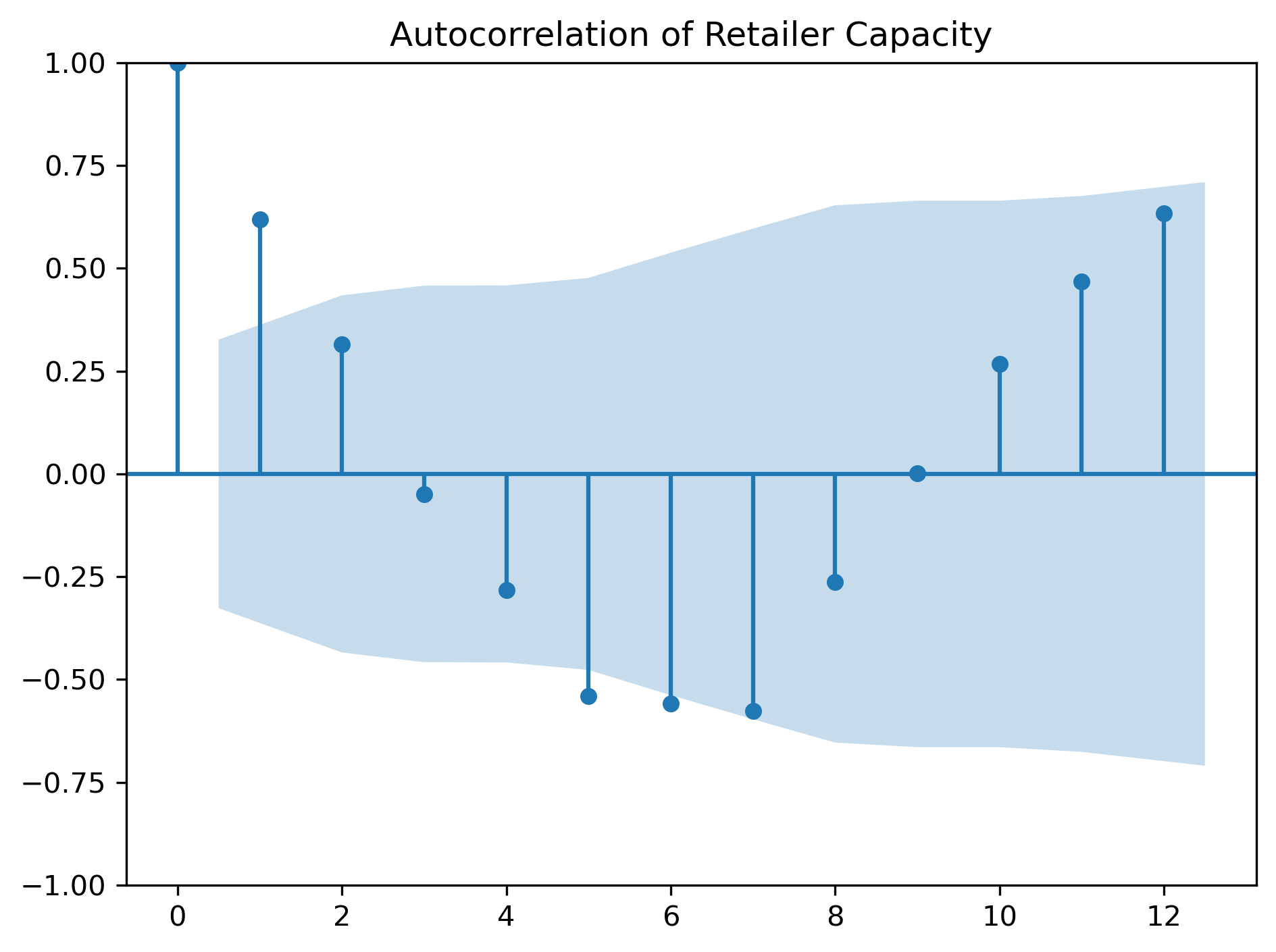}
    \caption{Autocorrelation of Rate of Return}
    \label{fig:autocorrelation_rate_of_return}
\end{figure}

Figures~\ref{fig:autocorrelation_retailer_capacity} and~\ref{fig:autocorrelation_rate_of_return} display the autocorrelation functions (ACF) for retailer capacity and rate of return, respectively. Autocorrelation measures the correlation of a time series with its own lagged values, offering insights into the persistence or randomness of the data. For retailer capacity (Figure~\ref{fig:autocorrelation_retailer_capacity}), the ACF shows a significant spike at lag 0, indicating perfect correlation with itself, as expected. However, the autocorrelation drops sharply and fluctuates around zero for lags 1 through 12, with values generally staying within the confidence intervals (shaded blue region). This suggests that retailer capacity exhibits little to no significant autocorrelation beyond the immediate time point, implying that past capacity values have minimal influence on future values over these lags. The presence of a few spikes outside the confidence interval (e.g., at lags 1 and 11) may indicate sporadic short-term dependencies or seasonal effects, but overall, the series appears to be largely random or weakly dependent over time.

Similarly, the ACF for the rate of return (Figure~\ref{fig:autocorrelation_rate_of_return}) shows a similar pattern, with a strong autocorrelation at lag 0 and rapid decay toward zero for subsequent lags. The values remain within the confidence intervals for most lags, with occasional spikes (e.g., at lags 1 and 11) suggesting minor short-term correlations. This indicates that the rate of return for Beer-G is also largely random or exhibits weak temporal dependence, with no strong long-term patterns detectable from the autocorrelation structure alone.

\subsection{Distribution of Numerical Variables}
\begin{figure}[h]
    \centering
    \includegraphics[width=8.5cm]{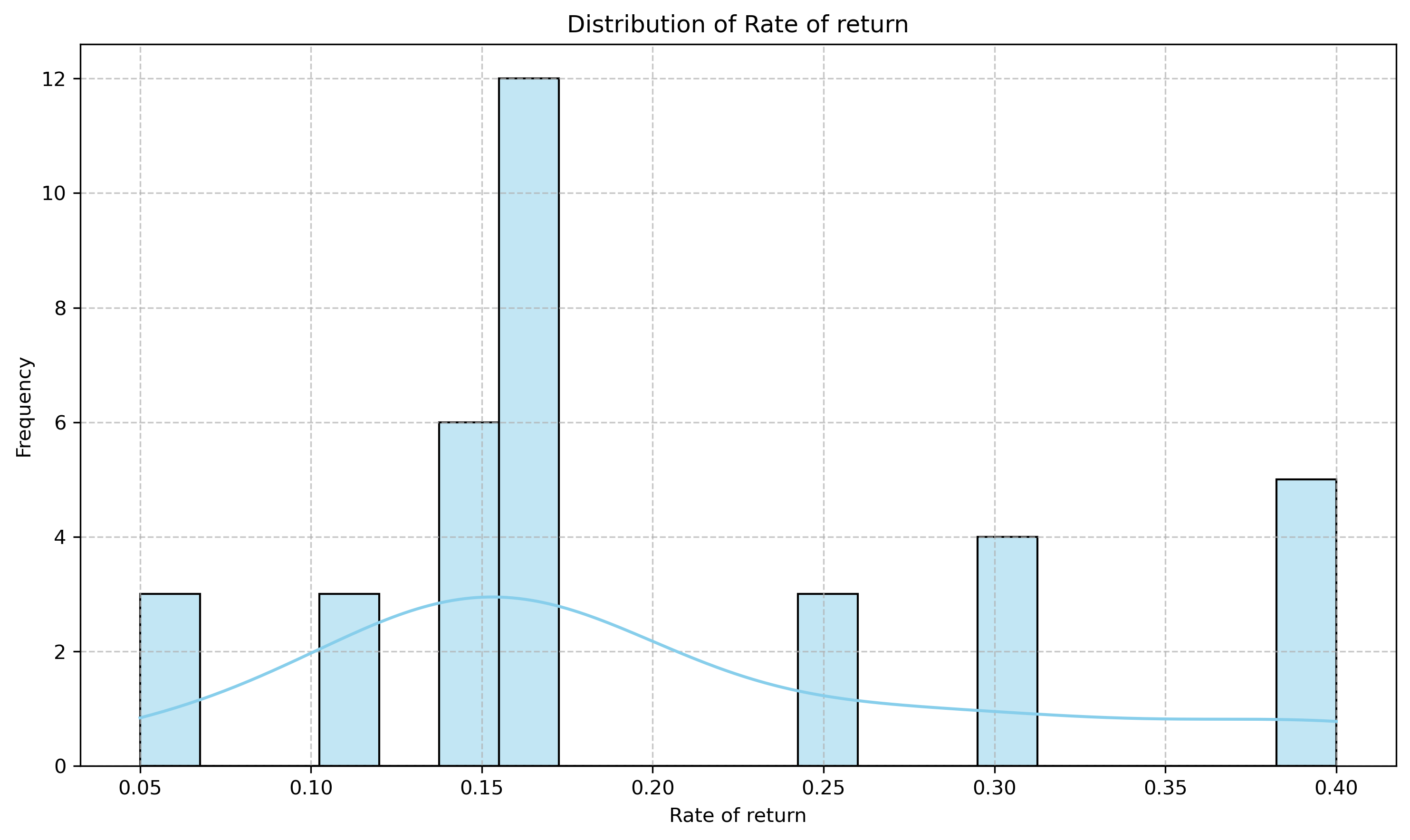}
    \caption{Distribution of Rate of Return}
    \label{fig:distribution_rate_of_return}
\end{figure}

\begin{figure}[h]
    \centering
    \includegraphics[width=8.5cm]{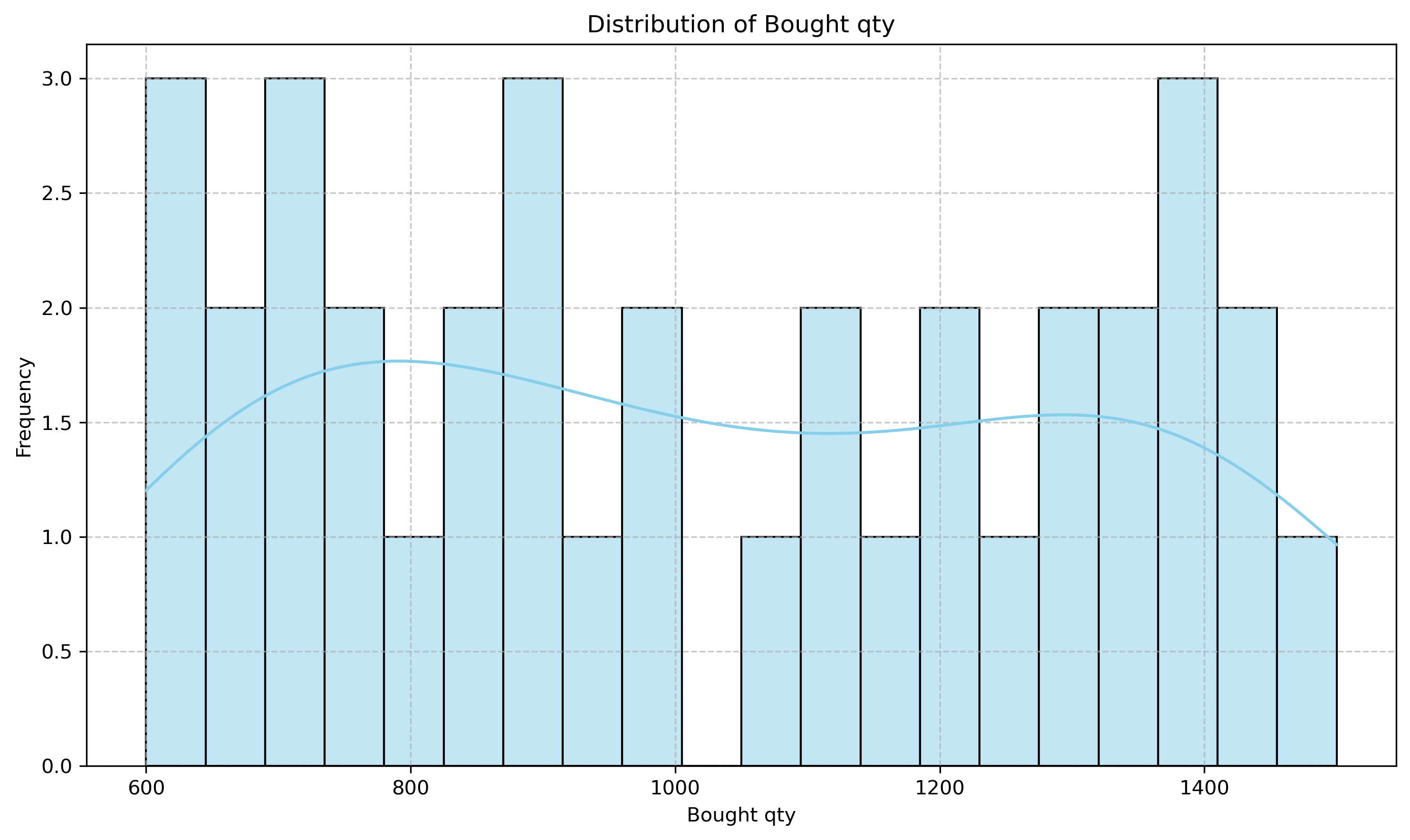}
    \caption{Distribution of Bought Qty}
    \label{fig:distribution_bought_qty}
\end{figure}

\begin{figure}[h]
    \centering
    \includegraphics[width=8.5cm]{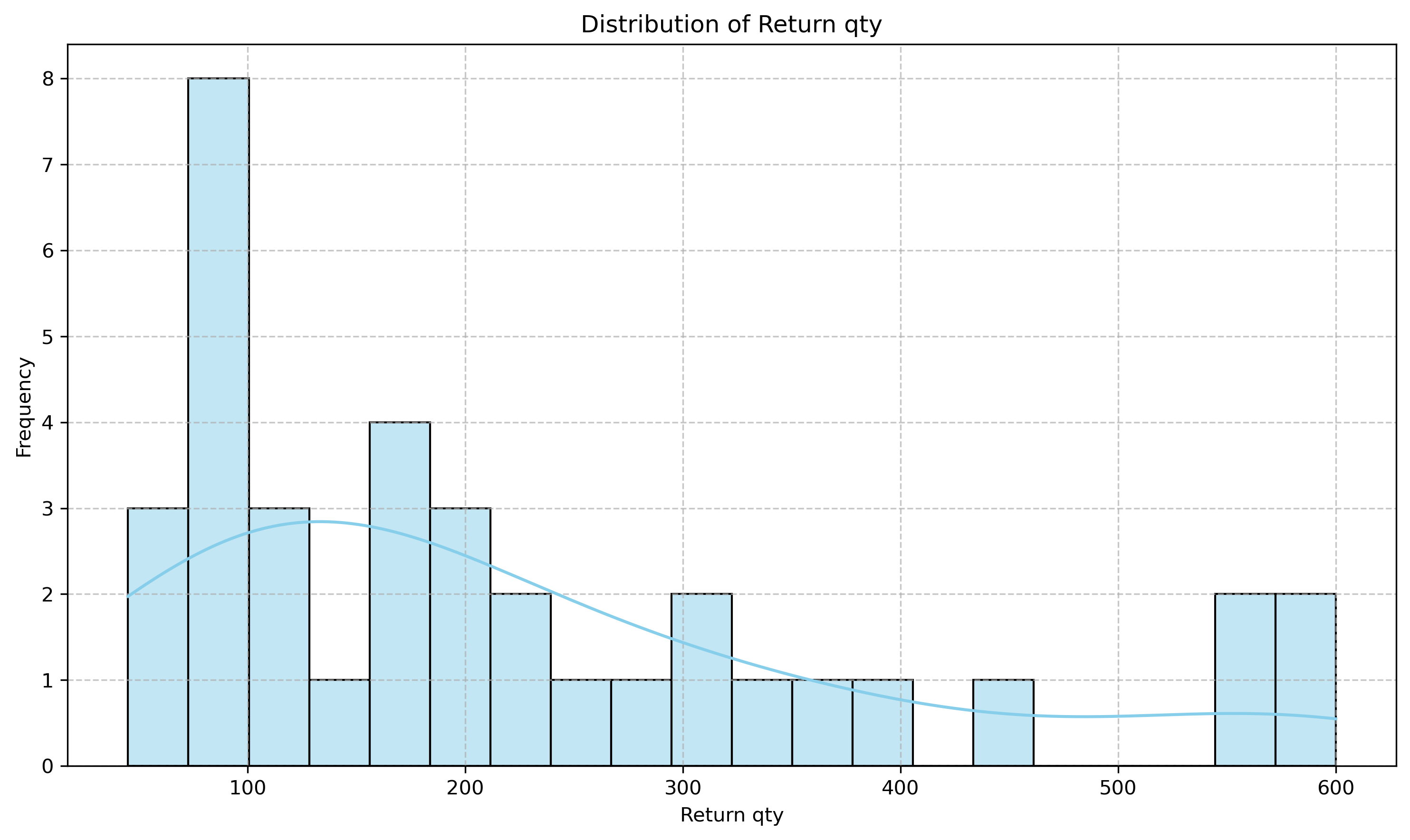}
    \caption{Distribution of Return Qty}
    \label{fig:distribution_return_qty}
\end{figure}

\begin{figure}[h]
    \centering
    \includegraphics[width=8.5cm]{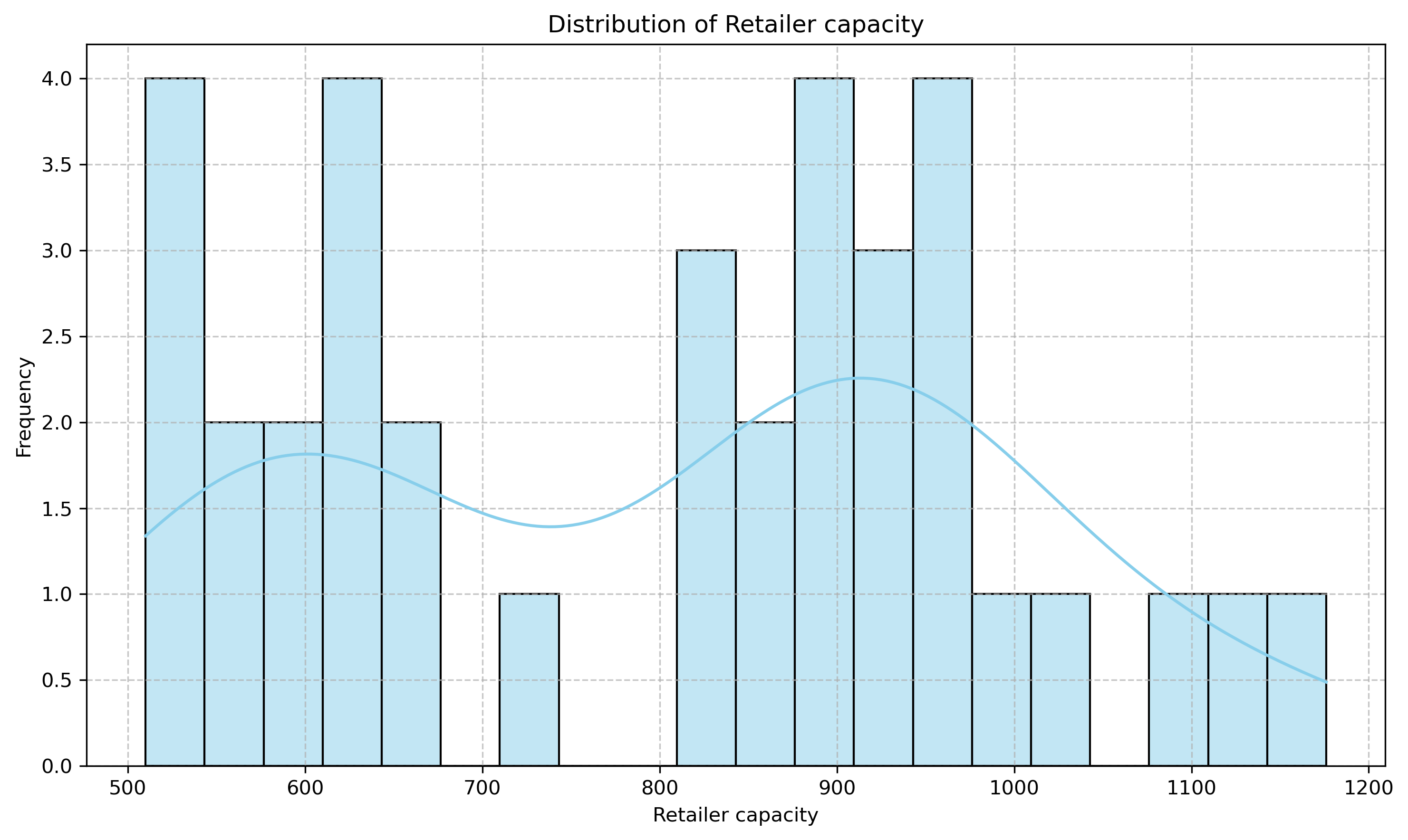}
    \caption{Distribution of Retailer Capacity}
    \label{fig:distribution_retailer_capacity}
\end{figure}

Figures~\ref{fig:distribution_rate_of_return},~\ref{fig:distribution_bought_qty},~\ref{fig:distribution_return_qty}, and~\ref{fig:distribution_retailer_capacity} present histograms of the distributions for `rate\_of\_return`, `bought\_qty`, `return\_qty`, and `retailer\_capacity`, respectively, providing a detailed view of their frequency distributions. 

- The distribution of `rate\_of\_return` (Figure~\ref{fig:distribution_rate_of_return}) shows a right-skewed distribution, with the highest frequency (approximately 12) occurring at a rate of return of 0.15, indicating that most return rates for Beer-G cluster around this value. Frequencies decrease as the rate of return increases beyond 0.15, with values of 0.05, 0.10, 0.20, 0.25, 0.30, 0.35, and 0.40 showing progressively lower frequencies, ranging from 2 to 5. The blue curve overlay suggests a unimodal distribution with a peak at 0.15, reflecting a typical return rate for the product.

- For `bought\_qty` (Figure~\ref{fig:distribution_bought_qty}), the histogram reveals a multimodal distribution, with peaks at 600, 700, 900, and 1400, each showing frequencies around 2.0 to 3.0. The distribution indicates that purchase quantities for Beer-G are spread across a range, with higher frequencies at lower quantities (600–900) and a notable peak at 1400, suggesting occasional large purchases. The blue curve overlay shows a wavy pattern, indicating potential seasonality or variability in buying behavior.

- The distribution of `return\_qty` (Figure~\ref{fig:distribution_return_qty}) is also right-skewed, with the highest frequency (approximately 8) at 100, indicating that most returns occur at low quantities. Frequencies decrease steadily as return quantities increase, with values of 200, 300, 400, 500, and 600 showing progressively lower frequencies (4, 2, 1, 1, and 2, respectively). The blue curve overlay confirms a unimodal distribution peaking at 100, reflecting a typical low return volume for Beer-G.

- For `retailer\_capacity` (Figure~\ref{fig:distribution_retailer_capacity}), the histogram shows a relatively uniform distribution with peaks at 600, 700, and 900, each with frequencies around 4.0, and lower frequencies (1.0) at 800, 1000, 1100, and 1200. This suggests that retailer capacity for Beer-G is distributed across a moderate range, with higher frequencies at lower to mid-capacity levels. The blue curve overlay indicates a multimodal pattern, reflecting variability in capacity allocation.

These histograms collectively highlight the variability and concentration of Beer-G’s key metrics, with `rate\_of\_return` and `return\_qty` showing right-skewed distributions, while `bought\_qty` and `retailer\_capacity` exhibit multimodal patterns, suggesting diverse purchasing and capacity behaviors.

\subsection{Relationship Between Rate of Return and Retailer Capacity}
\begin{figure}[h]
    \centering
    \includegraphics[width=8.5cm]{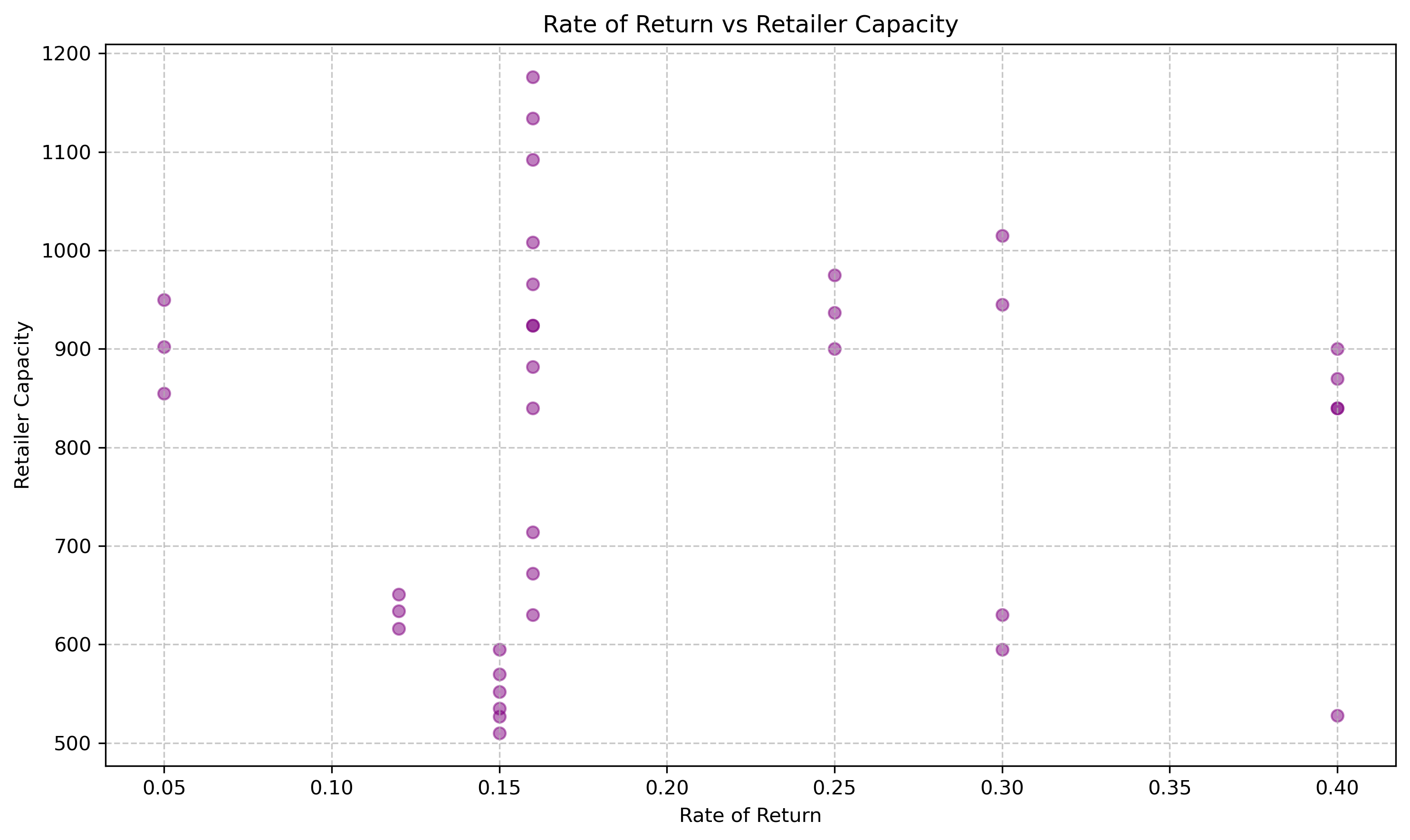}
    \caption{Rate of Return vs. Retailer Capacity}
    \label{fig:rate_of_return_vs_retailer_capacity}
\end{figure}

Figure~\ref{fig:rate_of_return_vs_retailer_capacity} presents a scatter plot of `rate\_of\_return` versus `retailer\_capacity`, providing insight into their relationship. The plot shows a cloud of purple points distributed across a range of `rate\_of\_return` values (0.05 to 0.40) and `retailer\_capacity` values (500 to 1200). There is no clear linear trend, but a general clustering suggests that higher retailer capacities (800–1200) are associated with a broader range of return rates, particularly around 0.15–0.25. Lower capacities (500–700) tend to have lower return rates, possibly indicating that constrained capacity limits returns. The lack of a strong correlation (consistent with the correlation heatmap showing a coefficient of 0.04) suggests that retailer capacity and rate of return are largely independent, with other factors (e.g., product quality, demand) driving return rates.

\subsection{Historical Time Series Analysis}
\begin{figure}[h]
    \centering
    \includegraphics[width=8.5cm]{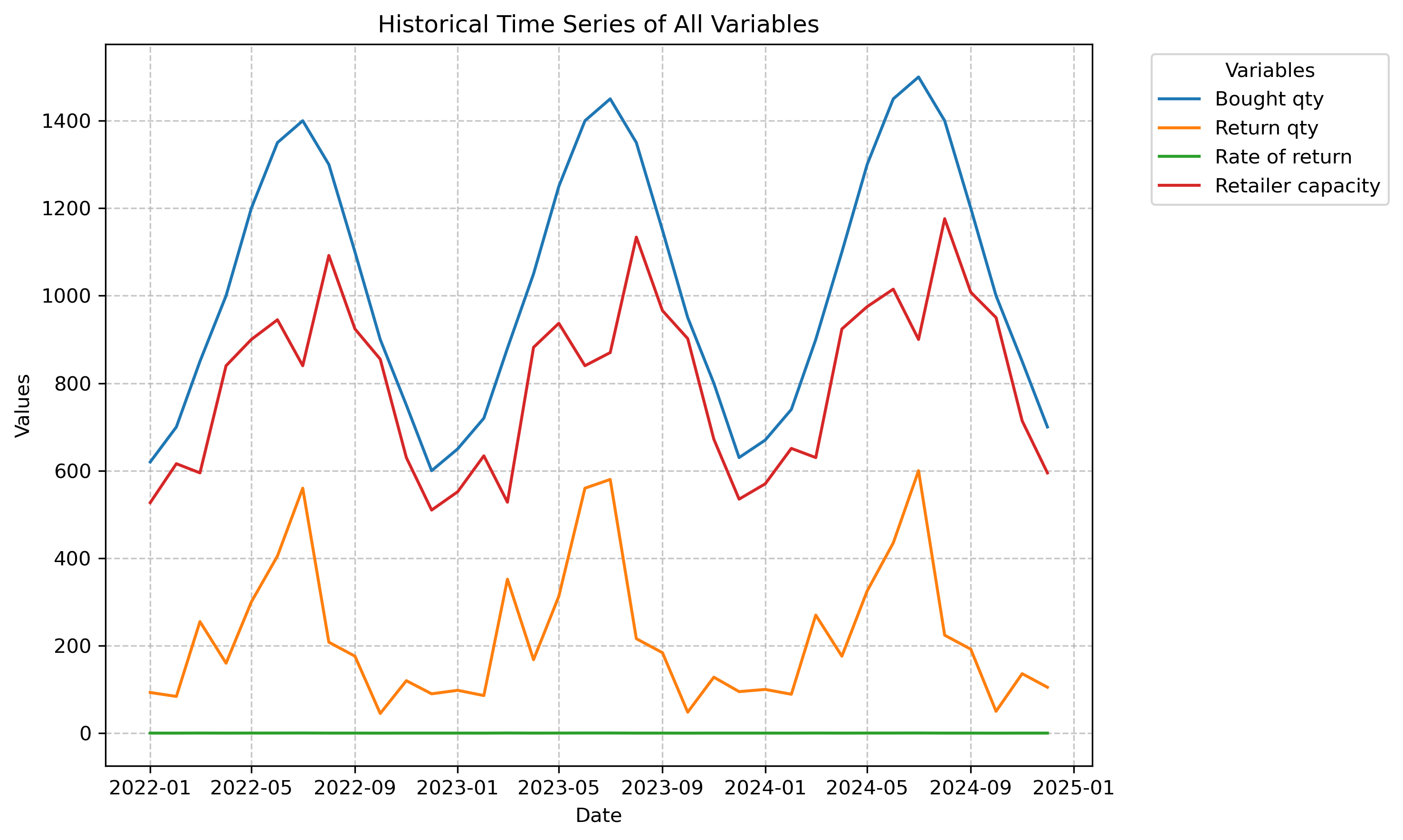}
    \caption{Historical Time Series of All Variables}
    \label{fig:historical_time_series_all_variables}
\end{figure}

Figure~\ref{fig:historical_time_series_all_variables} displays the historical time series of all variables (`bought\_qty`, `return\_qty`, `rate\_of\_return`, and `retailer\_capacity`) from January 2022 to January 2025. The plot uses different colors to distinguish each variable: `bought\_qty` (blue), `return\_qty` (orange), `rate\_of\_return` (green), and `retailer\_capacity` (red). 

- `Bought\_qty` (blue) shows significant volatility, with peaks reaching 1400 and troughs around 600, indicating large fluctuations in purchase volumes over time. Peaks occur around early 2022, mid-2023, and late 2024, suggesting seasonal or demand-driven surges.
- `Return\_qty` (orange) follows a similar but lower-amplitude pattern, peaking around 600 and dropping to near 0, indicating that returns are closely tied to purchases but at a much smaller scale.
- `Rate\_of\_return` (green) remains relatively stable, hovering around 0.2–0.4, with minor fluctuations, reflecting a consistent return rate over time despite purchase variability.
- `Retailer\_capacity` (red) mirrors `bought\_qty` to some extent, with peaks around 1200 and troughs around 600, but shows a declining trend after mid-2024, possibly due to capacity constraints or reduced demand.

This time series highlights the interdependence of these variables, with `bought\_qty` and `retailer\_capacity` showing the highest volatility, while `rate\_of\_return` remains relatively stable, and `return\_qty` scales with purchases but at a lower magnitude.

\subsection{Historical and Forecasted Trends}
\begin{figure}[h]
    \centering
    \includegraphics[width=8.5cm]{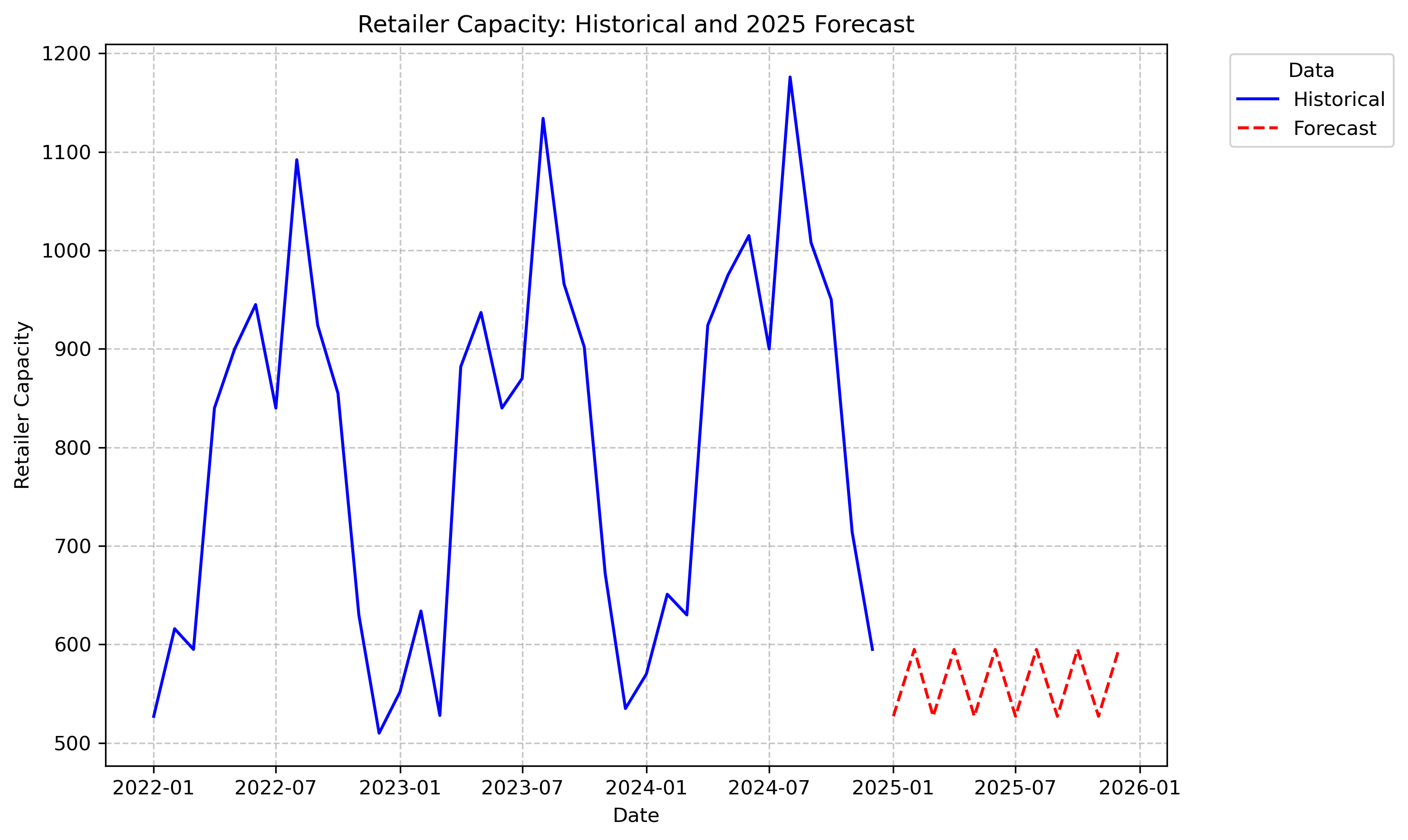}
    \caption{Retailer Capacity: Historical and 2025 Forecast}
    \label{fig:retailer_capacity_historical_forecast}
\end{figure}

\begin{figure}[h]
    \centering
    \includegraphics[width=8.5cm]{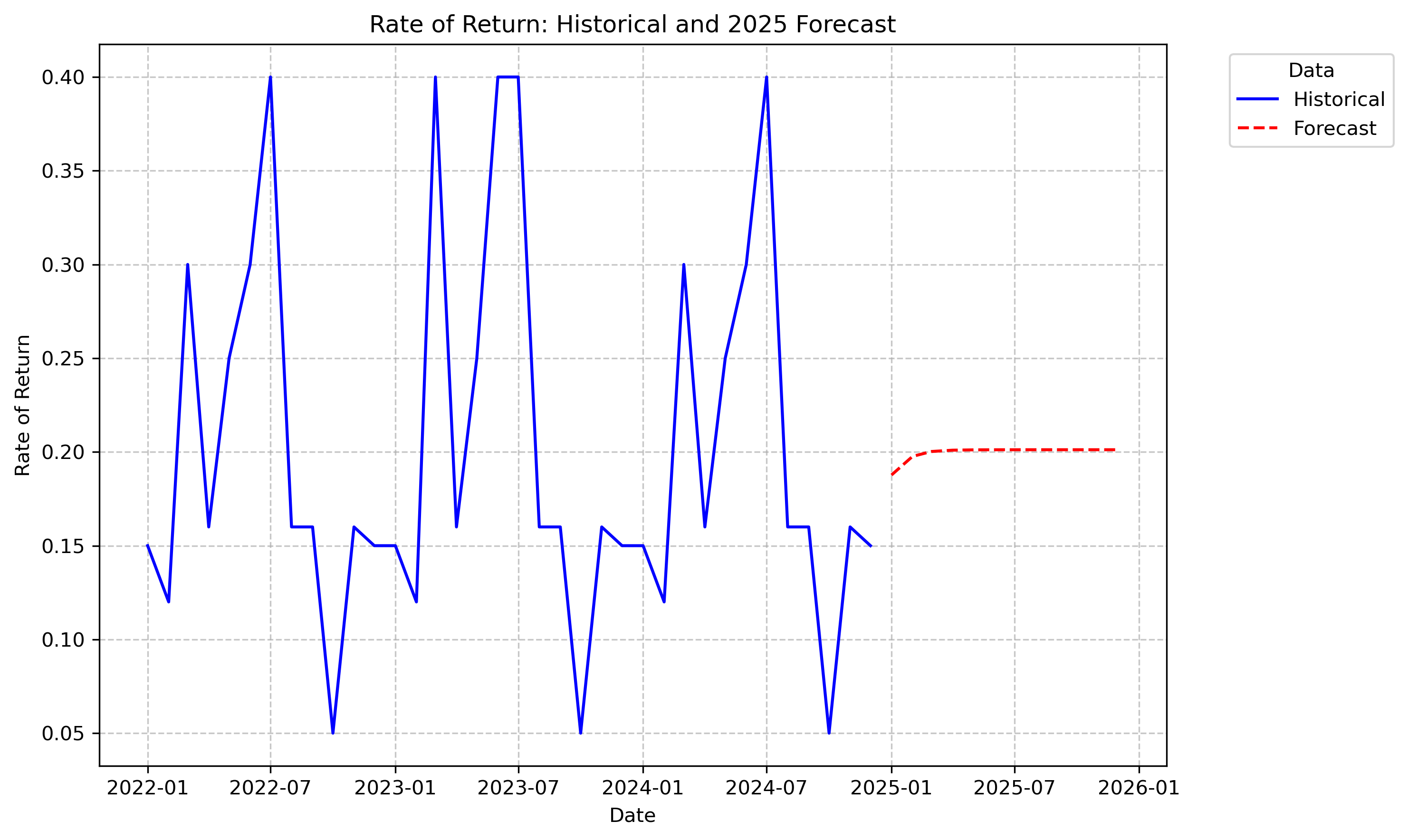}
    \caption{Rate of Return: Historical and 2025 Forecast}
    \label{fig:rate_of_return_historical_forecast}
\end{figure}

Figures~\ref{fig:retailer_capacity_historical_forecast} and~\ref{fig:rate_of_return_historical_forecast} illustrate the historical data and 2025 forecasts for retailer capacity and rate of return, respectively, covering the period from January 2022 to January 2026. For retailer capacity (Figure~\ref{fig:retailer_capacity_historical_forecast}), the historical data (blue line) shows significant volatility, with peaks reaching approximately 1,200 and troughs dropping to around 500 between 2022 and early 2025. This volatility suggests fluctuating demand or supply constraints for Beer-G over this period. The forecast for 2025–2026 (red dashed line) predicts a stabilization at a lower level, oscillating around 600–700, indicating a potential decline or stabilization in retailer capacity in the near future. This forecasted trend may reflect anticipated market saturation, reduced demand, or operational adjustments by retailers.

For the rate of return (Figure~\ref{fig:rate_of_return_historical_forecast}), the historical data (blue line) exhibits high volatility, with values ranging between 0.05 and 0.40 from 2022 to early 2025. Peaks occur periodically, suggesting intermittent high return rates, possibly due to quality issues, seasonal returns, or customer dissatisfaction with Beer-G. The 2025–2026 forecast (red dashed line) predicts a stable rate of return around 0.20, indicating a potential normalization or improvement in product quality or customer satisfaction, reducing return rates in the future.

\subsection{Correlation Analysis}
\begin{figure}[h]
    \centering
    \includegraphics[width=8.5cm]{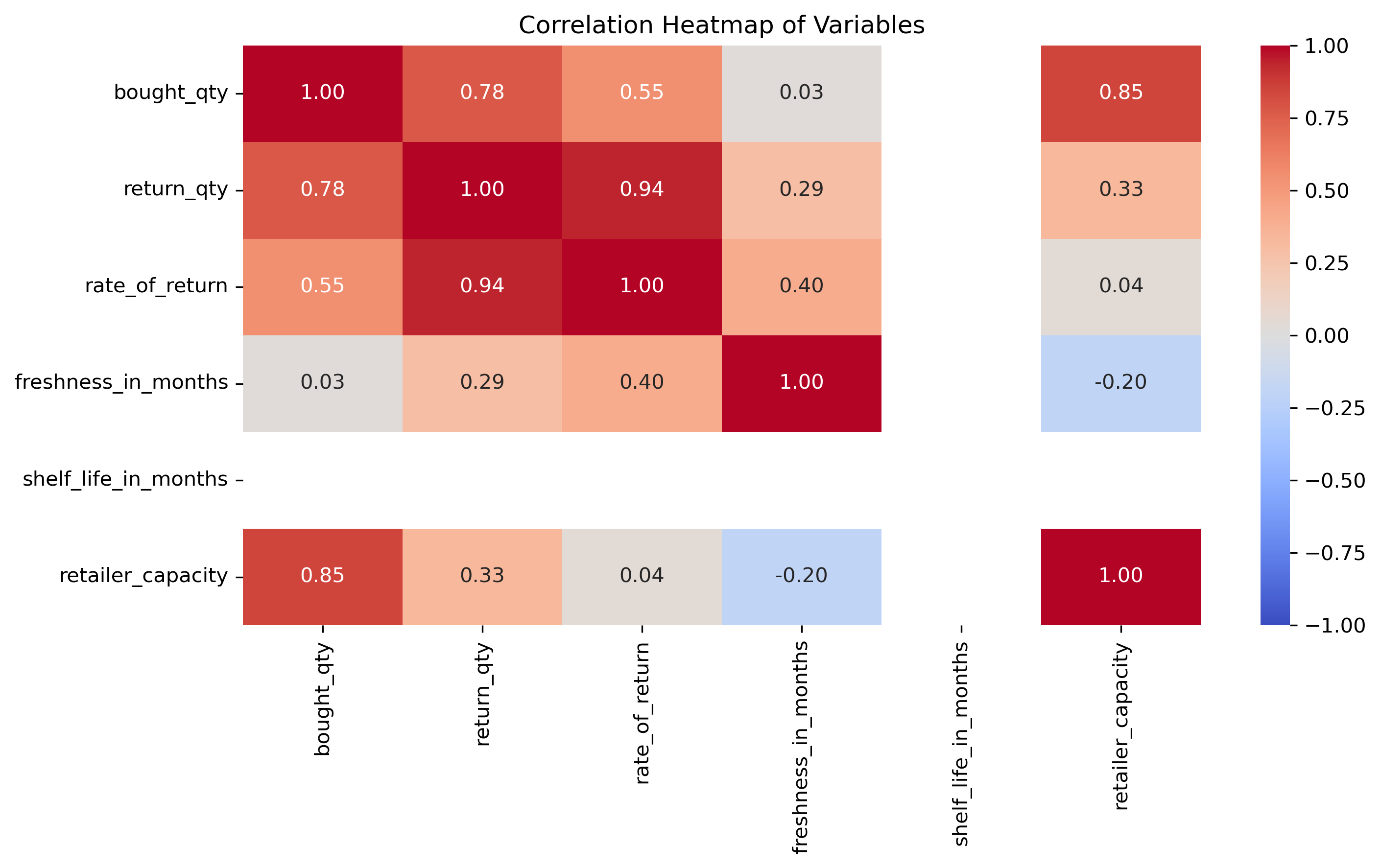}
    \caption{Correlation Heatmap of Variables}
    \label{fig:correlation_heatmap_variables}
\end{figure}

Figure~\ref{fig:correlation_heatmap_variables} presents a correlation heatmap of the variables, providing insight into their interrelationships. The heatmap reveals strong positive correlations among `bought\_qty`, `return\_qty`, and `rate\_of\_return`, with correlation coefficients of 1.00, 0.78, and 0.94, respectively, between these pairs. This suggests that higher purchase quantities are closely associated with higher return quantities and rates, potentially indicating quality control challenges or customer dissatisfaction with Beer-G. `Retailer\_capacity` shows a strong positive correlation with `bought\_qty` (0.85) and a moderate positive correlation with `return\_qty` (0.33), but a negligible correlation with `rate\_of\_return` (0.04) and a weak negative correlation with `freshness\_in\_months` (-0.20) and `shelf\_life\_in\_months` (-0.20). The negative correlations with freshness and shelf life may suggest that longer shelf life or freshness reduces the need for high retailer capacity, possibly due to slower turnover or spoilage concerns for Beer-G.

\subsection{Implications for Beer-G Product}
The statistical analysis shows Beer-G exhibits notable volatility in retailer capacity, purchases, and returns, with weak autocorrelation indicating short-term randomness rather than sustained trends. A strong purchase-return correlation suggests potential quality or satisfaction issues, meriting further exploration into defects or marketing discrepancies.
\begin{table*}[ht!]
\caption{Predicting Organic Beer-G 1 Liter Bad Risk Score for 2025}
\begin{center}
\begin{tabular}{|c|c|c|c|c|c|c|c|c|}
\hline
\textbf{Month} & \textbf{Demand Plan} & \textbf{Return Rate} & \textbf{Expected Return} & \textbf{Retailer Inventory} & \textbf{Freshness Left} & \textbf{Shelf Life} & \textbf{Freshness } & \textbf{BG Risk } \\
 & \textbf{(Sales Qty)} & \textbf{(\%)} & \textbf{(Qty)} & \textbf{Capacity (Monthly)} & \textbf{(Months)} & \textbf{(Months)} &\textbf{Ratio} &\textbf{Score} \\
\hline\hline
Jan & 500 & 18.77 & 94 & 527 & 2 & 4 & 0.5 & 0.422 \\
\hline
Feb & 600 & 19.76 & 119 & 595 & 1 & 4 & 0.25 & 0.668 \\
\hline
Mar & 700 & 20.02 & 140 & 527 & 3 & 4 & 0.75 & 0.37 \\
\hline
Apr & 800 & 20.09 & 161 & 595 & 0 & 4 & 0.0 & 1 \\
\hline
May & 900 & 20.11 & 181 & 527 & 4 & 4 & 1.0 & 0.343 \\
\hline
Jun & 1000 & 20.12 & 201 & 595 & 2 & 4 & 0.5 & 0.582 \\
\hline
Jul & 1200 & 20.12 & 241 & 527 & 1 & 4 & 0.25 & 0.823 \\
\hline
Aug & 1300 & 20.12 & 262 & 595 & 3 & 4 & 0.75 & 0.54 \\
\hline
Sep & 1400 & 20.12 & 282 & 527 & 0 & 4 & 0.0 & 1 \\
\hline
Oct & 1450 & 20.12 & 292 & 595 & 4 & 4 & 1.0 & 0.49 \\
\hline
Nov & 1500 & 20.12 & 302 & 527 & 2 & 4 & 0.5 & 0.757 \\
\hline
Dec & 1500 & 20.12 & 302 & 595 & 1 & 4 & 0.25 & 0.844 \\
\hline
\end{tabular}
\label{tab:beer_g_data}
\end{center}
\end{table*}

\subsection{Interpretation of Bad Goods Risk Scores}
The `BG Risk Score` in Table~\ref{tab:beer_g_data} categorizes the risk of Organic Beer-G 1 Liter becoming a bad good (e.g., defective, expired, or returned) into three levels: Low Risk (score $<$ 0.4), Medium Risk (0.4 $\le$  score $<$0.8), and High Risk (score $\ge$ 0.8) calculated using equation \ref{eq:return_rate}-\ref{eq:risk_score}. These levels indicate the likelihood of bad goods, with High Risk requiring immediate action, Medium Risk suggesting preventive measures, and Low Risk recommending monitoring. Based on this classification, we analyze the risk scores for each month and propose actionable recommendations to mitigate risks by adjusting `Demand Plan (Sales Qty)`, `Retailer Inventory Capacity (Monthly)`, or `Freshness Left (Months)`:

\begin{itemize}
    \item \textbf{January (0.422, Medium Risk)}: The Medium Risk score suggests a moderate likelihood of bad goods. Consider increasing `Freshness Left (Months)` from 2 to 3, or reducing `Demand Plan` to 450 if `Retailer Inventory Capacity (Monthly)` (527) is strained, to lower the risk to Low.
    
    \item \textbf{February (0.668, Medium Risk)}: The Medium Risk score indicates a moderate likelihood of bad goods. Increase `Freshness Left (Months)` from 1 to 2 or 3, or reduce `Demand Plan` to 550 if `Retailer Inventory Capacity (Monthly)` (595) is constrained, to mitigate the risk to Low.
    
    \item \textbf{March (0.37, Low Risk)}: The Low Risk score indicates minimal risk. No immediate action is required, but consistent monitoring of `Freshness Left` (currently 3 months) is advised to prevent spoilage.
    
    \item \textbf{April (1.0, High Risk)}: The High Risk score indicates a critical likelihood of bad goods. Adjust `Demand Plan` to 700 (reduce by 100) to align with `Retailer Inventory Capacity (Monthly)` (595), increase `Freshness Left` to 2, and ensure capacity matches demand to reduce the risk to Medium or Low.
    
    \item \textbf{May (0.343, Low Risk)}: The Low Risk score suggests minimal risk. No immediate action is needed, but maintain the high `Freshness Left` of 4 months to ensure product quality.
    
    \item \textbf{June (0.582, Medium Risk)}: The Medium Risk score indicates a moderate likelihood of bad goods. Increase `Freshness Left` to 3 or reduce `Demand Plan` to 900 if `Retailer Inventory Capacity (Monthly)` (595) is insufficient, to lower the risk to Low.
    
    \item \textbf{July (0.823, High Risk)}: The High Risk score suggests a critical likelihood of bad goods. Reduce `Demand Plan` to 1000, increase `Freshness Left` to 2, and ensure `Retailer Inventory Capacity (Monthly)` (527) supports demand to mitigate the risk to Medium or Low.
    
    \item \textbf{August (0.54, Medium Risk)}: The Medium Risk score indicates a moderate likelihood of bad goods. Increase `Freshness Left` to 4 or closely monitor inventory to prevent spoilage and reduce the risk to Low.
    
    \item \textbf{September (1.0, High Risk)}: The High Risk score indicates a critical likelihood of bad goods. Adjust `Demand Plan` to 1200 (reduce by 200), increase `Freshness Left` to 2, and ensure `Retailer Inventory Capacity (Monthly)` (527) is sufficient to lower the risk to Medium or Low.
    
    \item \textbf{October (0.49, Low Risk)}: The Low Risk score suggests minimal risk. Monitor `Freshness Left` and capacity, but no immediate action is needed.
    
    \item \textbf{November (0.757, Medium Risk)}: The Medium Risk score indicates a moderate likelihood of bad goods. Reduce `Demand Plan` to 1300, increase `Freshness Left` to 3, and adjust `Retailer Inventory Capacity (Monthly)` (527) to match demand to lower the risk to Low.
    
    \item \textbf{December (0.844, High Risk)}: The High Risk score suggests a critical likelihood of bad goods. Reduce `Demand Plan` to 1300, increase `Freshness Left` to 2, and ensure `Retailer Inventory Capacity (Monthly)` (595) supports demand to mitigate the risk to Medium or Low.
\end{itemize}
These recommendations aim to balance demand, capacity, and freshness to minimize the risk of bad goods, particularly for months classified as Medium or High Risk, ensure product quality, reduce returns, and optimize supply chain operations. 


\section{COMPARISON OF OUR WORK WITH PRESENT STATE-OF-THE-ART TECHNIQUES}\label{sec:compare}

Time series analysis has become a vital tool for predicting risks associated with defective or "bad" goods in supply chain management, enabling the forecasting of trends and the evaluation of risks based on historical data. Researchers have employed various time series models—such as ARIMA, Exponential Smoothing, Holt-Winters, and LSTM (Long Short-Term Memory)—to address supply-chain risk prediction 
\begin{table*}[hbt!]
\caption{Quantitative comparison of state-of-the-art bad goods risk prediction techniques}
\begin{center}
\begin{tabular}{ |c|c|c|c|c|c|c|c| }
 \hline
 \textbf{No} & \textbf{Author} & \textbf{Time Series Model} & \textbf{Risk Assessment} & \textbf{Machine Learning/Statistical} & \textbf{Dataset Used} & \textbf{Prediction} & \textbf{Risk Scoring} \\
 \hline
 \hline
 1 & Hyndman et al.~\cite{hyndman2002} & Exponential Smoothing & \ding{55} & \textbf{Statistical} & Simulated & \ding{51} & \ding{55} \\ \hline
 2 & Taylor~\cite{taylor2003} & Holt-Winters & \ding{55} & \textbf{Statistical} & Retail Sales & \ding{51} & \ding{55} \\ \hline
 3 & Zhang et al.~\cite{zhang2017} & LSTM & \ding{51} & \textbf{Deep Learning} & Supply Chain Data & \ding{51} & \ding{51} \\ \hline
 4 & Billah et al.~\cite{billah2006} & ARIMA & \ding{55} & \textbf{Statistical} & Economic Data & \ding{51} & \ding{55} \\ \hline
 5 & Proposed Work & ARIMA & \ding{51} & \textbf{Time Series ARIMA} & \textbf{Collected} & \ding{51} & \ding{51} \\
 \hline
\end{tabular}
\label{tab:com}
\end{center}
\end{table*}

challenges. In this section, we compare our proposed approach with existing state-of-the-art techniques, as summarized in Table~\ref{tab:com}.

Our method integrates Time Series ARIMA with a robust risk assessment framework to predict bad goods risks and assign risk scores, setting it apart from techniques that focus solely on forecasting or employ alternative statistical and machine learning approaches. For instance, Hyndman et al.~\cite{hyndman2002} utilize Exponential Smoothing for forecasting but do not incorporate risk assessment, limiting its utility for proactive risk management. Similarly, Taylor~\cite{taylor2003} applies the Holt-Winters method to retail sales data for prediction, yet it lacks a risk scoring component. In contrast, Zhang et al.~\cite{zhang2017} leverage LSTM for both prediction and risk assessment in supply chain contexts, though their deep learning approach is computationally intensive and less interpretable than our ARIMA-based solution. Billah et al.~\cite{billah2006} employ ARIMA for forecasting economic time series but do not extend it to risk scoring, unlike our work.

Our proposed approach, by combining ARIMA with risk scoring and applying it to a comprehensive collected dataset, offers a computationally efficient, interpretable, and effective solution for bad goods risk prediction in supply chains. This dual focus on prediction and risk assessment enhances decision-making capabilities for supply chain stakeholders, positioning our method as a valuable contribution to the field.

\section{CONCLUSIONS AND FUTURE WORK}\label{sec:future}
The proposed Time Series ARIMA-based predictive analytics model for calculating bad goods risk scores demonstrates significant potential in enhancing supply chain risk management for Organic Beer-G 1 Liter. By integrating historical data, time series forecasting, and a structured risk assessment framework, our approach accurately predicts bad goods risks, categorizing them into Low, Medium, and High Risk levels, and provides actionable recommendations to mitigate these risks. The experimental results, as shown in Table~\ref{tab:beer_g_data}, highlight the model’s effectiveness in identifying critical risk periods and optimizing demand, capacity, and freshness to minimize bad goods occurrences.

Future work could explore the following directions:
\begin{itemize}
    \item \textbf{Advanced Machine Learning Models}: Investigate the integration of deep learning models, such as LSTM or GRU, with ARIMA to improve prediction accuracy and handle non-linear patterns in bad goods risks.
    \item \textbf{Real-Time Data Integration}: Extend the model to incorporate real-time supply chain data, enabling dynamic risk scoring and immediate response to emerging risks.
    \item \textbf{Expanded Dataset}: Include additional variables (e.g., environmental factors, transportation conditions) and larger, diverse datasets to enhance the robustness and generalizability of the risk prediction model.
    \item \textbf{Automated Decision Support}: Develop an automated decision support system that leverages the risk scores to provide real-time recommendations for supply chain stakeholders, improving operational efficiency.
\end{itemize}

These advancements could further strengthen the model’s applicability and impact in supply chain risk management, addressing the evolving challenges of bad goods in the beverage industry and beyond.

\bibliographystyle{unsrt}

\end{document}